\pdfoutput=1

\documentclass[11pt]{article}

\usepackage[final]{acl}

\usepackage{times}
\usepackage{latexsym}

\usepackage[T1]{fontenc}

\usepackage[utf8]{inputenc}

\usepackage{microtype}

\usepackage{inconsolata}

\usepackage{graphicx}
\usepackage{todonotes}

\usepackage{afterpage}
\usepackage{pdflscape}

\usepackage{tcolorbox}
\usepackage{colortbl}

\usepackage{xspace}
\usepackage{tikz}

\title{From Disagreement to Understanding:\\ The Case for Ambiguity Detection in NLI}

\author{
\textbf{Chathuri Jayaweera} \and
\textbf{Bonnie J. Dorr}
\\
University of Florida, Gainesville, FL,USA \\
\texttt{\{chathuri.jayawee, bonniejdorr\}@ufl.edu}}

\begin{document}
\maketitle
\begin{abstract}
This position paper argues that annotation disagreement in
Natural Language Inference (NLI) is not mere noise but often reflects meaningful 
variation, especially when triggered by ambiguity in the premise or hypothesis. 
While 
underspecified guidelines and annotator behavior 
contribute to variation, content-based ambiguity provides a process-independent signal of 
divergent human perspectives. We call for a shift toward ambiguity-aware NLI 
that first identifies ambiguous input pairs, classifies their types,
and only then proceeds to inference.
To support this shift, we present a framework that incorporates ambiguity detection and classification prior to inference.
We also introduce a unified taxonomy that synthesizes
existing taxonomies,
illustrates key 
subtypes with examples,
and motivates 
targeted detection methods that better align models with human interpretation.
Although current resources lack datasets explicitly
annotated for ambiguity and 
subtypes, this gap presents an opportunity:
by developing new annotated resources and 
exploring unsupervised approaches to ambiguity detection, we 
enable more robust, explainable, and human-aligned NLI systems.
\end{abstract}

\section{Introduction}

This paper takes a position on how disagreement in Natural Language Inference (NLI) is best understood and modeled. While prior work has often treated annotator disagreement as noise---something to be minimized or resolved \cite{snow_cheap_2008, bowman_large_2015}---we argue that such disagreement can reflect meaningful, coexisting interpretations grounded in linguistic ambiguity. 

NLI, also known as Recognizing Textual Entailment (RTE) \cite{dagan_pascal_2005},
aims to classify the relationship between a premise (P) and a hypothesis (H). Suppose P1$=$\textit{John likes Mary}, P2$=$\textit{John lives near Mary}, H1$=$\textit{John knows Mary}, H2$=$\textit{John doesn't know Mary}. Standard inference labels would assign \textit{entailment} to (P1, H1), \textit{contradiction} to (P1, H2), and \textit{neutral} to (P2, H1). However,  humans may diverge: (P1, H2) could also be neutral if \textit{likes} is interpreted as distant admiration (e.g., of a celebrity).

\begin{figure}[t]
    \centering
    \small
    \input{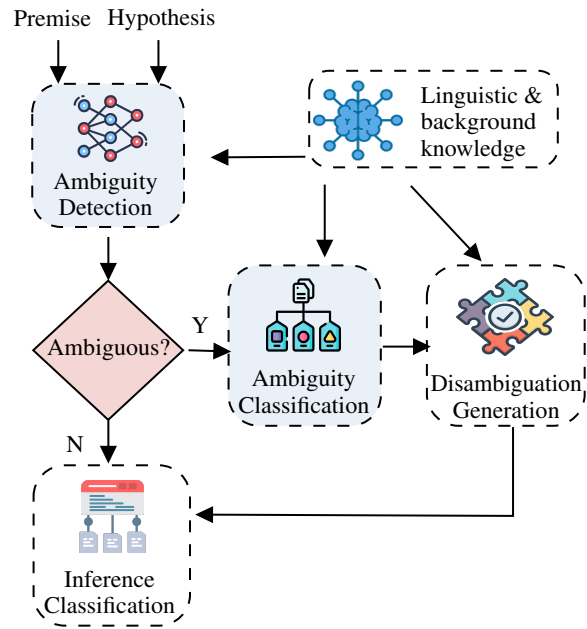}
    \vspace*{-.15in}
\caption{Framework 
for ambiguity-aware NLI:
First detect
whether a (P)remise or (H)ypothesis is ambiguous and, if so, classify the ambiguity type. 
Generate disambiguated versions and pass these to the inference classifier. 
Linguistic and other relevant background knowledge inform 
each 
stage. Gray $=$ focus of this paper; white $=$ supporting stages.}
\label{fig:proposed_pipeline}
\vspace*{-.2in}
\end{figure}

We adopt a perspectivist reframing of NLI that treats variation in inference classifications not as a flaw but as an inherent feature of natural language understanding. 
We emphasize content-based ambiguity as a central source of disagreement and advocate for deeper exploration of its role in shaping inference judgments. 

We situate our analysis within a framework for handling ambiguous NLI instances (Figure~\ref{fig:proposed_pipeline}).
This framework first determines whether an input pair is ambiguous; if so, it disambiguates the pair into distinct yet plausible human interpretations, enabling 
predictions aligned with each interpretation.


NLI is pivotal in understanding semantic relationships and 
is central to evaluating how well language models 
process
natural language.
NLI benchmarks are 
typically constructed
using human-annotated
entailment labels
\cite{bowman_large_2015, williams_broad-coverage_2018}. 
Despite frequent disagreements among annotators on the ``correct'' label for a given premise-hypothesis pair, most
NLI research 
assumes a 
single
``true'' inference for each 
case. 

Instances that deviate
from this 
assumption
are either filtered out 
during dataset construction \cite{bayer_mitres_2005} or handled 
via
majority vote \cite{bowman_large_2015}, 
based on the belief that
annotation disagreements 
reflect random
error rather than
systematic variation.
However, this approach contradicts the original
purpose of
NLI: to model what a 
reasonable, attentive, and informed human 
would plausibly infer from 
text \cite{manning_local_2006}. 

Recent studies have challenged the assumption that annotation disagreements
are mere noise,
demonstrating instead that such disagreements exhibit reproducible patterns grounded in legitimate interpretive differences \cite{pavlick_inherent_2019}. 
This recognition has motivated efforts to model the full distribution of plausible human inferences
\cite{chen_uncertain_2020, meissner_embracing_2021}. 

While these efforts are important, understanding \textit{why} such differences arise is equally critical for developing systems that reflect multiple human interpretations.
This position paper argues for an NLI modeling goal that centers on \textbf{identifying and categorizing ambiguity into recurring interpretive patterns, rather than merely modeling annotator distributions to capture coexisting human perspectives}.
We support this position through a review and analysis of existing research.  

The next section reviews 
work on modeling annotator label distributions and their limitations, 
motivating a closer look at sources of disagreement in NLI. 
Section~\ref{sec:disagreement-sources} examines prior 
categorizations of these sources,
Section~\ref{sec:significance-of-ambiguity} highlights
the unique role of ambiguity in premise-hypothesis pairs,
and Section~\ref{sec:understanding-ambiguity-in-nli} surveys current 
ambiguity-focused NLI research 
and future directions.

\section{Modeling annotator distribution}
Most NLI benchmarks are constructed using human-annotated entailment labels, which 
often result in cases where multiple annotators assign different labels to the same premise-hypothesis pair. 
From the early days of NLI research, scholars have expressed
concerns
about how to handle such disagreements
\cite{bayer_mitres_2005}.

Re-annotation of
the RTE1 development and training sets reveals substantial discrepancies between the 
original
and new 
labels \cite{bayer_mitres_2005}.
Even after filtering out problematic examples, human judges only achieve a 
91\% agreement rate.
Similar disagreements in the Stanford Natural Language Inference (SNLI)
dataset 
complicate the process of learning robust
decision boundaries for each entailment label \cite{pan_discourse_2018}.

Such
cases are typically 
treated as ``annotation noise,'' 
resolved by assigning 
a majority label 
under the assumption that 
one 
``true'' inference exists for each premise-hypothesis pair. However, growing evidence suggests that these
disagreements reflect systematic, reproducible variation rather than random error
\cite{pavlick_inherent_2019}.
In many cases, divergent annotations signal the existence of multiple
plausible 
interpretations.

Current
NLI models, trained on
majority-labeled benchmarks,
struggle to capture the full distribution of human judgments and tend to perform better 
when annotator agreement is high \cite{nie_what_2020}. This 
highlights both a dependence on 
agreement 
and a failure to model collective human reasoning.
\citet{meissner_embracing_2021} further
show that models trained on 
soft labels---distributions over annotator responses---better approximate human judgments and improve single-label prediction accuracy.

These findings have inspired a 
growing line of research focused on 
modeling human opinion
distributions.
For example, the 
Uncertain Natural Language Inference (UNLI) framework \cite{chen_uncertain_2020} proposes predicting 
subjective 
probabilities of entailment rather than 
coarse categorical labels. 
While UNLI captures
a more probabilistic notion of inference,
it targets average responses and does not attempt to model the full range of interpretations.

\citet{zhang_identifying_2021} 
contrast
systematic inference (high agreement) 
and ambiguous cases (high disagreement).
They 
build artificial
annotators 
using
BERT
\cite{devlin_bert_2019} 
to simulate annotation variation,
enabling downstream models to 
determine if
a given premise-hypothesis
pair is likely to elicit disagreement.
\citet{zhou_distributed_2022} further improve modeling of opinion
distributions 
beyond standard softmax 
assumptions.

Together, these studies mark a shift from a prescriptive view---assuming a single correct label---toward a descriptive approach that acknowledges interpretive variations.
They 
help pave the way for
systems that capture 
ambiguity inherent 
in
natural language. 
However, simply modeling disagreement alone does not explain \textit{why} interpretations diverge.
To advance beyond descriptive modeling, 
we argue that NLI systems must also 
systematically identify and categorize the sources of disagreement---especially content-based ambiguity---as a foundation for more perspective-sensitive inference \cite{plank_problem_2022}. 
We next examine the 
sources that give rise to divergent judgments in NLI.

\section{Disagreement sources in NLI}
\label{sec:disagreement-sources}

\begin{figure}[ht]
    \centering
    \small
    \vspace*{.1in}
    \input{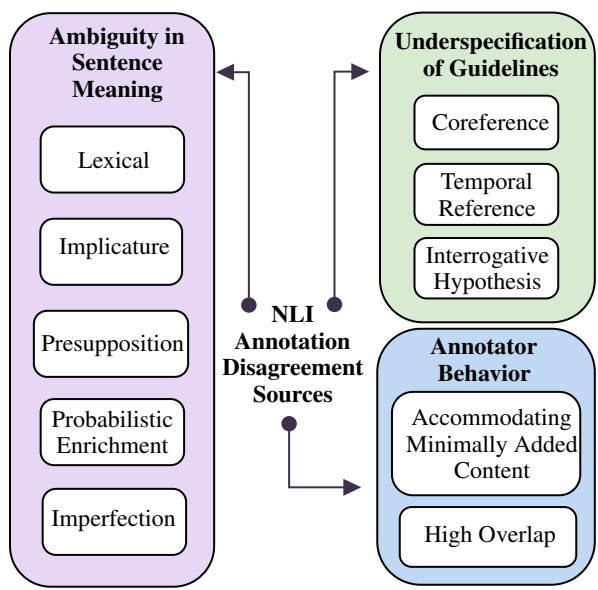}
    \vspace*{-.2in}
    \caption{The taxonomy of disagreement sources 
    developed by \citet{jiang_investigating_2022}, 
    building on \citet{aroyo_truth_2015}.
    While these frameworks classify sources of disagreement, we argue for \textit{reframing} such disagreement as a signal of coexisting interpretations to 
    model---not noise to be resolved.    }
\label{fig:disagreement-taxonomy}
\end{figure}

According to the “Triangle of Reference” \cite{aroyo_truth_2015},
disagreement in 
annotation arises from three main sources: 
(1) interpretative 
ambiguity in the
\textit{input content} itself 
(Uncertainty in sentence meaning); 
(2) unclear
annotation guidelines (Underspecification in guidelines); and (3) differences in annotators’
background knowledge or task understanding 
(Annotator behavior). 
This 
framework maps directly 
onto annotation workflows 
in NLI benchmarks. 
Building on this foundation,
\citet{jiang_investigating_2022} propose 
a more fine-grained taxonomy 
for NLI, refining each 
category into subtypes that 
reflect recurring premise-hypothesis patterns 
(Figure~\ref{fig:disagreement-taxonomy}). 

In the subsections below, 
we adopt a variant of this taxonomy, reframing \textit{Uncertainty in Sentence Meaning} as \textit{Ambiguity in Sentence Meaning} (Section~\ref{sec:ambiguity-sentence-meaning}), a shift already noted in Figure~\ref{fig:disagreement-taxonomy}. This distinction is central to our position. For completeness, we also briefly describe the roles of guideline underspecification (Section~\ref{sec:guideline-underspec}) and annotator behavior (Section~\ref{sec:annotator-behavior}), though these are not the central emphasis of our position. 

In addition, while we view Jiang and de Marneffe's taxonomy as a valuable classification framework, we go further: rather
than treating disagreement as noise
to be explained or resolved,
we reframe it as a meaningful signal of coexisting interpretations---something to be modeled directly as part of the NLI task. 



\subsection{Ambiguity in Sentence Meaning} 
\label{sec:ambiguity-sentence-meaning}

Ambiguity in sentence meaning---manifesting as multiple plausible interpretations---is a major source of disagreement in NLI annotations.
In the taxonomy 
introduced by \citet{jiang_investigating_2022}, this form of content-based ambiguity 
is further divided into five subtypes: Lexical, Implicature, Presupposition, Probabilistic Enrichment, and Imperfection.
Together, these categories reflect
the range of
interpretative uncertainty
that 
arises from the language content itself, independent of annotator knowledge or instructions specified in annotation guidelines.

\texttt{Lexical} arises when a word or phrase in the premise or hypothesis has multiple possible senses or is underspecified.
\texttt{Implicature}
refers to 
cases where
the hypothesis expresses a logical or pragmatic implication of the premise, 
leaving room for divergent judgments depending on the reader's 
perspective. \texttt{Presupposition}
covers instances where the hypothesis draws on background presuppositions introduced by the premise, which may or may not be universally shared.
\texttt{Probabilistic} \texttt{Enrichment}
denotes cases where the inference relationship is not categorical, but depends on plausibility or likelihood, producing variation in 
individual perception (Figure~\ref{fig:probabilistic-enrichment}). 
\texttt{Imperfection}
includes 
typos, grammatical errors, or fragmented phrasing that
impede clear interpretation.

\begin{figure}[ht]
\begin{footnotesize}
\vspace*{.1in}
\begin{tabular}{|p{0.97\linewidth}|}\hline
\arrayrulecolor{black}
\rowcolor{violet!20} 
\textbf{Probabilistic Enrichment} \\ \hline
\textcolor{blue}{\textbf{Premise:}} “I think this report shows that we have had an inordinately productive and successful year.”\\
\textcolor{orange!80!black}{\textbf{Hypothesis:}} “The report shows that we need to be productive to have a successful year” \\ \hline
\end{tabular}
\end{footnotesize}
\caption{Probabilistic Enrichment ambiguity: Annotator's label choice---\texttt{Entailment} or \texttt{Neutral}---depends on whether the relationship between productivity and success mentioned in the premise is considered plausible or not.}
\label{fig:probabilistic-enrichment}
\vspace*{-.15in}
\end{figure}

While this categorization is based on a manually analyzed sample and shaped by the judgments of linguistically trained annotators, it offers a valuable foundation for surfacing and organizing patterns of interpretative variation in NLI.
Although it does not capture the full range of ambiguities present in premise-hypothesis pairs,
it provides an important starting point for tracing the roots of disagreement in NLI annotations and for recognizing how such divergences arise from legitimate differences in interpretation rather than annotation error.

\subsection{Underspecification
of Guidelines}
\label{sec:guideline-underspec}

Guideline underspecification is another source of annotation disagreement, but unlike content-based ambiguity, it often reflects task design flaws that can be addressed through clearer instructions.
Even when the premise-hypothesis pairs are unambiguous, annotators may diverge in how they interpret or apply the labeling instructions if those instructions lack sufficient precision or fail to address edge cases. \citet{jiang_investigating_2022} identify
three specific subtypes under this category: Coreference, Temporal Reference, and Interrogative Hypothesis.

\begin{figure}[ht]
\begin{footnotesize}
\vspace*{.12in}
\begin{tabular}{|p{0.97\linewidth}|}\hline
\arrayrulecolor{black}
\rowcolor{green!20} 
\textbf{Temporal Reference} \\ \hline
\textcolor{blue}{\textbf{Premise:}} “You wake up one bright autumn morning and you’re halfway to the subway when you decide to walk to work instead.”\\
\textcolor{orange!80!black}{\textbf{Hypothesis:}} “You wake up early and decide to walk instead of take the subway.” \\ \hline
\end{tabular}
\end{footnotesize}
\caption{Temporal Reference disagreement: Annotator's label choice---\texttt{Contradiction} or \texttt{Entailment}---depends on whether the decision is interpreted as happening before or after the commute, respectively. }
\label{fig:temporal-reference}
\end{figure}

\texttt{Coreference} cases
involve unclear assumptions about whether entities in the premise and hypothesis refer to the same thing.
Without guidance on how strongly to assume shared reference, annotators may reach inconsistent labels.

\texttt{Temporal} \texttt{Reference} arises when it
is unclear when the hypothesis should be evaluated.
In Figure~\ref{fig:temporal-reference} one annotator might interpret the decision as occurring \textbf{before} the commute (\texttt{Contradiction}), while another 
may align it with a decision made \textbf{during} the commute (\texttt{Entailment}).



The \texttt{Interrogative}
\texttt{Hypothesis}
covers cases where the hypothesis is phrased as a question.
Since questions are not truth-apt (i.e., 
not directly true or false), annotators 
must infer an implied assertion.
\citet{jiang_investigating_2022} 
focus
only on interrogative hypotheses, while others (e.g., 
\citet{gubelmann_when_2023}) argue that interrogative premises can 
cause similar confusion.

These sources of disagreement are worth recognizing, but they stem from instruction gaps rather than genuine interpretive variation---and thus lie outside this paper’s primary focus.

\subsection{Annotator Behavior}
\label{sec:annotator-behavior}
The third pillar of disagreement, according to the Triangle of Reference, 
is variation in annotation behavior: differences in background knowledge, beliefs, attention, or task interpretation across annotators. While often treated as noise in NLI pipelines, such variation can reflect meaningful differences in how people reason with language. Two annotators may bring different contextual assumptions to the same premise-hypothesis pair, leading to divergent but reasonable judgments.

\citet{jiang_investigating_2022} identify two specific behavioral tendencies that contribute to such disagreement:
\texttt{Accommodating} \texttt{Minimally} \texttt{Added} \texttt{Content} and
\texttt{High} \texttt{Overlap}.
The first involves hypotheses that add a small amount of plausible but unstated information. Some annotators accept this as implied, while others reject it based on stricter entailment criteria.
The second reflects a tendency to judge \texttt{Entailment} based on surface-level similarity---lexical or structural.
This can lead some to overestimate entailment based on form rather than meaning, while others focus on more subtle semantic distinctions (Figure~\ref{fig:high-overlap}).

\begin{figure}[ht]
\begin{footnotesize}
\vspace{.1in}
\begin{tabular}{|p{0.97\linewidth}|}\hline
\arrayrulecolor{black}
\rowcolor{blue!20} 
\textbf{High Overlap} \\ \hline
\textcolor{blue}{\textbf{Premise:}} “The sunlight, piercing through the branches, turned the auburn of her hair to quivering gold.”\\
\textcolor{orange!80!black}{\textbf{Hypothesis:}} “The auburn of her hair became golden then the sunlight hit it.” \\ \hline
\end{tabular}
\end{footnotesize}
    \caption{High Overlap disagreement: Annotators may infer \texttt{Entailment} between the premise and hypothesis due to the high lexical overlap, while the meaning of the two sentences suggests \texttt{Contradiction}.}
\label{fig:high-overlap}
\vspace*{-.1in}
\end{figure}

Some 
annotation tendencies stem not from errors, but from genuine interpretive variation. For example, \texttt{Accommodating} \texttt{Minimally} \texttt{Added} \texttt{Content} reflects meaningful differences, whereas \texttt{High} \texttt{Overlap} more likely signals annotation error. This underscores the importance of evaluating annotator behavior carefully, rather than assuming that all variation reflects valid perspectives. 

Among the various sources,
content-based ambiguity is the most direct and reliable indicator of genuine interpretive divergence. 
While understanding annotator behavior is useful, our focus is on
detecting ambiguity 
in the language itself.
Even so, recognizing 
behavioral patterns can inform future 
perspectivist NLI systems that accommodate multiple interpretations.
Next, we motivate why content-based ambiguity merits special attention relative to guideline and annotator effects.

\section{Why Does Content-Based Ambiguity Deserve Special 
Attention?} 
\label{sec:significance-of-ambiguity}

Among the disagreement sources outlined above, content-based ambiguity stands out as the only type that can be systematically addressed through computational modeling without relying on additional information about 
the annotators or the guidelines they follow. This form of ambiguity originates from the text itself, independent of the annotation process, yet it remains a fundamental driver of divergent interpretations. As such, it represents a root cause of disagreement inherent to natural language, posing a persistent challenge for inference systems aiming for consistent and reliable predictions.

\begin{figure}[ht]
\begin{footnotesize}
\vspace*{.12in}
\begin{tabular}{|p{0.97\linewidth}|}\hline
\arrayrulecolor{black}
\rowcolor{violet!20} 
\textbf{Implicature Ambiguity} \\ \hline
\textcolor{blue}{\textbf{Premise:}} “It hopes to bring on another 25 or 35 people when the new building opens next fall.”\\
\textcolor{orange!80!black}{\textbf{Hypothesis:}} “They already have a waiting list for the new building” \\ \hline
\end{tabular}
\end{footnotesize}
    \vspace*{-.05in}
    \caption{Implicature Ambiguity: Annotators may infer a waiting list from \textit{hopes to bring on another 25 or 35 people}. If so, they label it \texttt{Entailment}; if not, they label \texttt{Neutral}.}
\label{fig:implicature-ambiguity}
\vspace{-.1in}
\end{figure}
Consider the 
\texttt{Implicature}
ambiguity in Figure~\ref{fig:implicature-ambiguity}.
Some annotators interpret \textit{hopes to bring on another 25 or 35 people} as implying a \textit{waiting list} and choose 
\texttt{Entailment}.
Others focus strictly on what is stated and select
\texttt{Neutral}.
This interpretative variability stems 
from linguistic ambiguity rather than 
annotator background or faulty instructions.  
Such cases underscore the importance of treating
content-based ambiguity 
as central to analysis, rather than dismissed
as noise.

\citet{jiang_investigating_2022}’s findings indicate that the most common sources of disagreement fall under content-based ambiguity, underscoring its
prevalence. 
In contrast, issues related to underspecified guidelines can typically be resolved through clearer instructions and better annotation practices, meaning they do not strongly reflect genuine differences in human interpretation.

Similarly, while some annotator behaviors---such as 
\texttt{Accommodating} \texttt{Minimally} \texttt{Added} \texttt{Content}---reflect natural variation, others like 
\texttt{High} \texttt{Overlap}
may undermine the goals of the NLI task. Disagreements stemming from annotator behavior or guideline underspecification 
therefore warrant scrutiny 
before being treated as meaningful.
Not
all disagreements are noise, 
though some clearly reflect error
\cite{weber-genzel_varierr_2024}. In contrast, content-based ambiguity arises from the language itself and requires no external filtering or supervision, making it a uniquely reliable source of interpretive variation. Identifying such ambiguity supports the creation of disambiguated versions, allowing NLI benchmarks to better capture the range of plausible interpretations. Beyond benchmarking, ambiguity-aware modeling has practical consequences in downstream settings. 

Identifying content-based ambiguity in NLI data has significant real-world implications. NLI frequently serves as a core component of fact-verification pipelines, where it is used to assess the relationship between claims and supporting evidence \cite{thorne_fever_2018, jayaweera_amrex_2024}. Effectively pinpointing potential ambiguities---including those intentionally introduced---strengthens such pipelines by improving their capacity to flag potentially misleading content in real-world settings \cite{liu_were_2023}.

However, there are currently no established methods for disentangling disagreements caused by content-based ambiguity from those arising due to underspecified annotation guidelines or annotator behavior. As a result, most existing work focuses primarily on detecting premise-hypothesis pairs with high annotation disagreement, rather than investigating the underlying types of disagreement, particularly those stemming from ambiguity. Therefore, there is a necessity to build models that: (1) identify ambiguous premise-hypothesis pairs and (2) classify the respective ambiguity type.
These observations motivate two concrete tasks---ambiguity detection and ambiguity classification---which we discuss next.

\section{Understanding Ambiguity in NLI}
\label{sec:understanding-ambiguity-in-nli}
NLI systems aim to determine the inference relationship between a given premise and hypothesis, but ambiguity in either can complicate that process
(Figure ~\ref{fig:implicature-ambiguity}). This often leads to discrepancies among annotators, who may assign different inference labels based on their individual interpretations. 
In some cases, annotators may even agree on the same label while interpreting the text differently---a phenomenon known as within-label variation \cite{jiang_ecologically_2023}. Further complicating matters, 
ambiguity may arise
in the premise, the hypothesis, or both, 
increasing the complexity of inference decisions \cite{liu_were_2023}.

While some efforts have been made to develop models that detect instances with high annotator disagreement \cite{jiang_investigating_2022, jiang_ecologically_2023, park_where_2025}, there are no existing implementations that specifically identify or classify ambiguous instances in NLI—underscoring the need for systems designed to address this gap.

\subsection{Ambiguity Detection in NLI}

We define \textit{ambiguity detection} in NLI as identifying instances that 
elicit divergent 
interpretations due to input 
ambiguity---whether in the premise, the hypothesis, or both. 

\citet{jiang_investigating_2022} 
explore the detection of high-disagreement instances in NLI using multi-label prediction and a four-class classification scheme  
(\texttt{Entailment}, \texttt{Contradiction}, \texttt{Neutral}, and \texttt{Complicated}). However, their work does not go further to distinguish ambiguity as a specific cause of disagreement. \citet{jiang_ecologically_2023} build on this by incorporating explanations for disagreement but still focus solely on identifying highly contested instances.

\citet{park_where_2025} attempt to detect ambiguous cases in NLI benchmarks using hidden layer representations of Large Language Models (LLMs), but their training data includes disagreements from all categories, making the system 
a general disagreement detector rather than a model focused on ambiguity.
\citet{liu_were_2023} assess language models’ ability to detect ambiguous instances using the Ambient dataset, but their results show that model performance 
remains below human-level accuracy.

These studies reflect the current state of ambiguity detection in NLI, highlighting the need for further investigation. A key challenge in developing systems to identify ambiguous premise-hypothesis pairs is the lack of datasets annotated for ambiguity. Creating annotated datasets and exploring unsupervised methods are essential next steps. 

To address the current scarcity of ambiguity-type annotated NLI data, we leverage existing datasets that already incorporate disambigutations \cite{liu_were_2023} and explanations \cite{jiang_ecologically_2023} as assistive cues to annotate ambiguity types. This approach would help create a more cohesive dataset that integrates insights across the various taxonomies discussed in Section~\ref{sec:understanding-ambiguity-in-nli}.

At the same time, the limited scale of these resources highlights the need for additional strategies. Promising directions include data augmentation techniques such as paraphrasing, the continued use of manual annotation to ensure high-quality gold standards, and the strategic use of large language models (LLMs) as evaluators. Together, these methods can substantially expand the availability of annotated data, enabling both broader coverage of ambiguity types and more robust evaluation of ambiguity-aware NLI systems.

\subsection{Ambiguity Classification in NLI}

\textit{Ambiguity classification} 
identifies the exact type(s) of ambiguity present in a premise-hypothesis pair. Several taxonomies have been developed
to categorize the various forms of ambiguity found in NLI inputs.
As noted and illustrated in Figure~\ref{fig:disagreement-taxonomy}, \citet{jiang_investigating_2022} present a taxonomy comprising five ambiguity types. These have been identified in samples from the ChaosNLI \cite{nie_what_2020} and MNLI \cite{williams_broad-coverage_2018} datasets. 

\begin{figure}[ht]
    \centering
    \footnotesize
    \input{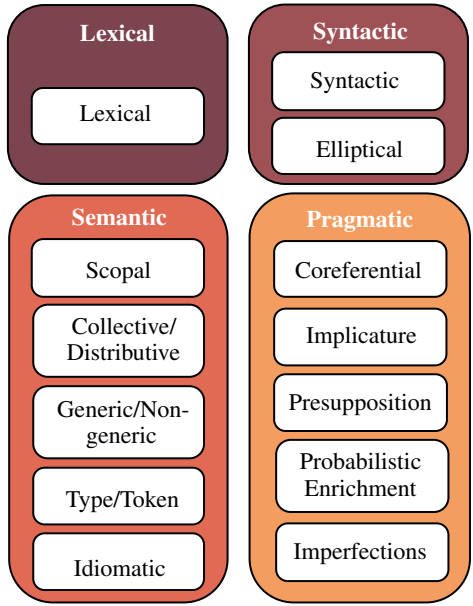}
    \caption{Unified ambiguity type taxonomy: We build on 
    prior taxonomies \cite{jiang_investigating_2022,liu_were_2023,li_taxonomy_2024}, organizing them into four broad types---Lexical, Syntactic, Semantic and Pragmatic---to support tailored detection strategies based on common characteristics.}
    \label{fig:unified-ambiguity-types}
    \vspace*{-.1in}
\end{figure}

\citet{liu_were_2023} introduce a 
taxonomy based on expert linguistic annotations of the Ambient dataset, which 
contains both curated and generated ambiguous premise-hypothesis pairs. 
They identify
additional ambiguity types
in NLI data, including \texttt{Syntactic}, \texttt{Pragmatic}, \texttt{Scopal}, and \texttt{Figurative} ambiguities, while grouping 
others under a residual
\texttt{Other} category.

Building on this, \citet{li_taxonomy_2024} 
refine the classification by proposing finer-grained 
types
such as Type/Token and 
Collective/Distributive, 
and aligning with 
\citet{jiang_investigating_2022} through the inclusion of
Presupposition and Implicature. These refinements reveal further
unexplored ambiguities that enhance the understanding of human interpretations.
Drawing on
these 
developments,
we present a unified taxonomy,\footnote{Refer to \cite{li_taxonomy_2024} for the definitions of each ambiguity type not described in this paper.} 
that organizes
ambiguity types into four broad categories---Lexical, Syntactic, Semantic, and Pragmatic---to support a more comprehensive 
view (Figure~\ref{fig:unified-ambiguity-types}).


However, to our knowledge, no existing system
automatically identifies 
the ambiguity type(s) present in a given premise–hypothesis pair. This gap highlights the practical importance of our framework: by organizing ambiguity types into linguistically grounded categories, we lay the foundation for developing detection methods tailored to each type. 
In doing so, we advance 
toward more nuanced, interpretable NLI models that 
not only detect
ambiguous input, but also explain \textit{how} and \textit{why} human interpretations 
diverge.

\section{Call to Action: Recognizing Ambiguity as 
Signal, Not Noise}
Disagreement among annotators in Natural Language Inference (NLI) is often treated as noise---something to minimize or discard. However, many of these disagreements reflect genuine interpretive differences, often triggered by ambiguity in the premise, hypothesis, or both. Our analysis suggests that while underspecified annotation guidelines and inconsistent annotator behavior can lead to label disagreement, such cases must be carefully scrutinized to distinguish between annotation errors and 
genuine
differences in human interpretation.

In contrast, ambiguity in the NLI input itself---whether lexical, syntactic, semantic, or pragmatic---serves as a clear, process-independent signal of interpretive variation, providing a basis for understanding how meaning can diverge across readers. This 
highlights 
a 
needed shift: from optimizing for annotator consensus to explicitly identifying and characterizing ambiguity as a central feature of natural language.

To support this shift, we outline the following two key directions:
\begin{itemize} 

\item\textbf{Identify Ambiguous Pairs:} Develop robust methods 
to detect
premise–hypothesis pairs that exhibit inherent ambiguity, using cues from linguistic theory, annotation patterns, and interpretability tools.

\item\textbf{Classify Ambiguity Types:}
Design strategies for distinguishing among different types of ambiguity.
A unified classification framework that groups ambiguity types based on their shared characteristics can offer a foundation for designing targeted identification methods.
\end{itemize}

A key novelty of this work lies in articulating a unified framework (Figure~\ref{fig:proposed_pipeline}) that extends beyond prior approaches focused solely on modeling annotation distribution. Whereas earlier efforts largely stop at detecting high-disagreement instances, this framework explicitly distinguishes ambiguity from other sources of variation by leveraging linguistic features and pertinent background knowledge.

The framework consists of four stages. The first
determines whether a premise-hypothesis pair is inherently ambiguous, thereby distinguishing between genuine interpretive variation 
from annotation noise and other sources of disagreement. If an instance is 
ambiguous, the second classifies the type(s),
creating systematic linkages to linguistic background knowledge.

The third stage 
generates relevant disambiguated versions, after which 
the fourth 
models inference classification 
for both ambiguous and non-ambiguous instances, based on predictions from 
earlier stages. The framework's strength lies in offering a structured and operational foundation for ambiguity-aware NLI, moving the field from descriptive accounts of disagreement toward a principled methodology that can be empirically tested.

By pursuing these goals, we can build NLI models that are not only more aligned with human interpretation, but also more explainable in predictions. Ambiguity-aware systems better align with human interpretation and produce more consistent, interpretable, and robust predictions.

This reframing is not only timely---it is essential for developing NLI systems that reflect the complexity of human understanding, rather than abstracting it away.

While we advocate for a shift toward ambiguity-aware NLI systems, realizing this vision is currently constrained by a key limitation: the lack of datasets that are explicitly annotated for ambiguity and categorized by ambiguity type. Most existing NLI datasets 
are not designed with interpretive variation or ambiguity classification in mind, making it difficult to systematically identify and analyze ambiguous instances or to evaluate models on their ability to handle them.

This gap limits the development and benchmarking of methods for 
detecting and classifying ambiguity. 
The absence of gold-standard annotations for different ambiguity types hinders progress in training and evaluating models that aim to align more closely with human interpretive processes.

To address this, we suggest two complementary directions. First, there is a clear need for the \textbf{creation of new datasets} specifically annotated for ambiguity presence and type. Such resources would lay the groundwork for both empirical analysis and model development. Second, we see promise in \textbf{exploring unsupervised or weakly supervised methods} that can surface potential ambiguities without requiring extensive manual labeling. Techniques leveraging patterns of annotator disagreement, discourse features, or model uncertainty could offer scalable alternatives in the absence of annotated data.

Despite current limitations, 
these strategies 
offer a promising path toward building NLI systems that better reflect the complexity and nuance of human language understanding.

\section*{Limitations}

The framework we articulate for ambiguity-aware NLI systems establishes a theoretical foundation, with empirical validation remaining an important next step. As a position paper, our aim is to 
stimulate discussion and motivate future empirical work. The framework's logical rigor and integration of existing taxonomies offer a strong basis
for future experimentation and evaluation.

Future research must address the creation of datasets explicitly annotated
for  ambiguity,
alongside the development and evaluation of systems 
to identify ambiguous instances in NLI. Such efforts will contribute to a deeper understanding by identifying indicators of different ambiguity types, and characterizing how they shape
inference judgments. 
We also anticipate exploring hybid approaches that combine
linguistic analysis with large language models 
to advance 
ambiguity detection for NLI.

\section*{Acknowledgments}

This work would not have been possible without the generous startup support provided by Dr. Herbert Wertheim through the Herbert Wertheim College of Engineering at the University of Florida.


\bibliography{latex/references}

\appendix



\end{document}